\documentclass[manuscript, acmtog]{acmart}
\usepackage{threeparttable}
\AtBeginDocument{%
  \providecommand\BibTeX{{%
    \normalfont B\kern-0.5em{\scshape i\kern-0.25em b}\kern-0.8em\TeX}}}

\copyrightyear{2021}
\acmYear{2021}
\setcopyright{acmlicensed}
\acmConference[EAAMO '21]{Equity and Access in Algorithms, Mechanisms, and Optimization}{October 5--9, 2021}{--, NY, USA}
\acmBooktitle{Equity and Access in Algorithms, Mechanisms, and Optimization (EAAMO '21), October 5--9, 2021, --, NY, USA}
\acmPrice{15.00}
\acmDOI{10.1145/3465416.3483299}
\acmISBN{978-1-4503-8553-4/21/10}

\begin{document}

\title{Mitigating Racial Biases in Toxic Language Detection with an Equity-Based Ensemble Framework}

\author{Matan Halevy}
\affiliation{%
  \institution{Georgia Institute of Technology}
  \city{Atlanta}
  \state{Georgia}
  \country{USA}}
\email{matan@gatech.edu}

\author{Camille Harris}
\affiliation{%
  \institution{Georgia Institute of Technology}
  \city{Atlanta}
  \state{Georgia}
  \country{USA}}
\email{charris320@gatech.edu}

\author{Amy Bruckman}
\affiliation{%
  \institution{Georgia Institute of Technology}
  \city{Atlanta}
  \state{Georgia}
  \country{USA}}
\email{asb@cc.gatech.edu}

\author{Diyi Yang}
\affiliation{%
  \institution{Georgia Institute of Technology}
  \city{Atlanta}
  \state{Georgia}
  \country{USA}}
\email{diyi.yang@cc.gatech.edu}

\author{Ayanna Howard}
\email{howard.1727@osu.edu}
\orcid{1234-5678-9012}
\affiliation{%
  \institution{The Ohio State University}
  \city{Columbus}
  \state{Ohio}
  \country{USA}
}

\renewcommand{\shortauthors}{Halevy et al.}

\begin{abstract}
Recent research has demonstrated how racial biases against users who write African American English exists in popular toxic language datasets. While previous work has focused on a single fairness criteria, we propose to use additional descriptive fairness metrics to better understand the source of these biases. We demonstrate that different benchmark classifiers, as well as two in-process bias-remediation techniques, propagate racial biases even in a larger corpus. We then propose a novel ensemble-framework that uses a specialized classifier that is fine-tuned to the African American English dialect. We show that our proposed framework substantially reduces the racial biases that the model learns from these datasets. We demonstrate how the ensemble framework improves fairness metrics across all sample datasets with minimal impact on the classification performance, and provide empirical evidence in its ability to unlearn the annotation biases towards authors who use African American English.

\noindent ** Please note that this work may contain examples of offensive words and phrases.
\end{abstract}



\maketitle
%%

% \maketitle{https://www.overleaf.com/project/60b84ac9720defc56cbc81f6}
\section{Introduction}
In response to the rise of hateful and toxic language used in online communities, its detection has become a growing field of interest for both researchers and industry professionals \cite{fortuna2018}. Social media companies have increased their automated moderation efforts in order to promote healthier discourse and reduce toxicity \cite{adl_2021}. However, the problem space is riddled with challenges and human biases that pose issues in both classifier performance and practical applications. Issues such as class imbalances, label subjectivity, and annotation biases can cause these algorithmic models to encompass and propagate human biases against the very minority groups they are designed to protect \cite{sap-etal-2019-risk,park-etal-2018-reducing,wich-etal-2020-impact}. 

The challenge of using machine learning systems to automate hateful language detection can be traced to a high-degree of subjectivity in the human-labeled datasets on which the algorithms are trained. These datasets rely on annotators' familiarity with cultural and historical contexts and ever-changing societal forms of bigotry \cite{fortuna2018}. Sap et al. \cite{sap-etal-2019-risk} documented the existence of annotation bias against users of African American English (AAE) in commonly used toxic language datasets. As a well-studied English dialect, AAE exhibits distinct grammatical rules and syntax and can be used as a proxy for racial identity when user race is not reported. The potential of falsely moderating AAE users comes at a time of increased online racial harassment towards African Americans \cite{adl_2021}. 

Existing literature (reviewed in \S\ref{related-works}) used false-positive rate (FPR), the probability of classifying non-toxic samples as toxic conditional on the samples being non-toxic, as the only criteria for correcting models' biases against AAE authors. We observe that AAE samples in popular toxic-language datasets are mainly annotated as toxic, leaving a very small sample of AAE instances that are true negatives (non-toxic). Hence, a fairness criterion based solely on FPR has a very limited scope.
 
In our work, we employ more descriptive fairness metrics (\S\ref{fairness-metrics-descr}), in addition to FPR, to evaluate how annotation biases against AAE authors propagate through hate detection models. (\S\ref{debiasing-algorithms-desc}) We then perform experiments on commonly used models and examine the results from two bias-mitigation strategies that have been proposed to reduce algorithmic bias towards group-identifiers  and disparities using FPR \cite{kennedy-etal-2020-contextualizing,prost-2019}. We demonstrate how all models continue to propagate and encompass racial biases from four toxic-language datasets in spite of the proposed bias-mitigation techniques. 

To address this issue, we propose an ensemble model architecture (\S\ref{ensemble-desc}) that uses a general toxic language classifier with a specialized AAE classifier, and show that it reduces the effects of annotation biases towards AAE users without impacting classifier performance. We then conduct error analysis on misclassified AAE tweets from this framework (\S\ref{error-analysis}) to better understand further challenges in debiasing and classifying AAE instances for toxic language detection.

\section{Datasets}
\subsection{Toxic Language Datasets}
Our work uses four publicly available toxic language datasets derived from Twitter that use compatible definitions of hate speech and toxic language. The four datasets are chosen due to their popularity in toxic language research and their corpus being sampled from Twitter. 

\subsubsection*{DWMW17 \cite{DavidsonWMW17}}
Davidson et al. randomly sampled 24,802 tweets that contained words and phrases from Hatebase.org. The tweets were annotated by 3 or more crowd-sourced annotators and assigned labels of: "Hate Speech", "Offensive", or "Neither". The definition for hate speech provided to the annotators was: “language that is used to expresses hatred towards a targeted group or is intended to be derogatory, to humiliate, or to insult the members of the group.” The annotators achieved 92\% intercoder agreement and the final label distribution were 77.4\% offensive language, 5.8\% hate, and 16.8\% were neither.

\subsubsection*{FDCL18 \cite{Founta2018}}
Founta et al. created a 79,996 sample dataset consisting of tweets annotated as either "Abusive, Hateful, Normal, or Spam." These tweets were collected from a stream of tweets and then filtered using sentiment analysis and phrases from Hatebase.org. The authors defined hate speech as: "language used to express hatred towards a targeted individual or group, or is intended to be derogatory, to humiliate, or to insult the members of the group, on the basis of attributes such as race, religion, ethnic origin, sexual orientation, disability, or gender." Inter-coder label agreement when holding out at most one of five annotators was 55.9\%. The final label distribution showed 66\% of the labels were normal, 16.8\% were spam, 12.6\% were abusive, and 4.5\% were hateful.

\subsubsection*{Golbeck \cite{golbeck18}}
Golbeck et al.'s binary dataset of tweets labeled as "Harassment" or "Not Harassment" was created from sampling tweets that contained hashtags and phrases that were present in their exploration of offensive tweets. They used two annotators in the first round and when there was disagreement, a third would be brought in to determine a majority label. The "Harassment" label ended up including sub-topics of racism, misogyny, homophobia, threats, hate speech, directed harassment. Non-harassment included potentially offensive and non-harassing tweets. Their definition of hate speech used was: “hate or extreme bias to a particular group. Could be based on religion, race, gender, sexual orientation, etc. Generally, these groups are defined by their inherent attributes, not by things they do or think.” The Harassment label accounted for 15.7\% of the tweets in the corpus. This dataset is only used in the \textit{Toxic} aggregate as it does not contain enough AAE samples individually. 

\subsubsection*{WH16 \cite{waseem-hovy-2016-hateful}}
The Waseem and Hovy (WH16) 16,849 sample dataset was collected by  sampling tweets containing at least one of the phrases or words they deemed to be hateful. The authors labeled the tweets as racist, sexist or neither using guidelines inspired by critical race theory and had a domain expert review their labels. However, this dataset has received significant criticism from scholars \cite{klubicka2018examining,schmidt-wiegand-2017-survey}, who deride it for most of the racist tweets being anti-muslim and the sexist tweets relating to a debate over an Australian television show. Additionally, this dataset can introduce author bias as it is noted that two users wrote 70\% of sexist tweets and 99\% of racist tweets were written by another single user. Due to these limitations, we only include the positive instances of this dataset in the aggregated datasets \textit{Toxic} and \textit{Hate} as its use in an aggregate largely addresses the author and topic biases. 

\subsection{Data Preparation}
\label{sec:data-prep}
To estimate the dialect of the tweet, we used Blodgett et al.'s \cite{blodgett-etal-2016-demographic} dialect estimation model that outputs the posterior probability of the text sample belonging to the dialect of AAE or Standard American English (SAE). Due to the low incidence of AAE tweets within the toxic language datasets, we used a lower threshold of $Pr(AAE \geq 0.6)$ relative to Blodgett et al.'s use of $Pr(AAE \geq 0.8)$ in order to have a larger sample of AAE tweets. In the \S\ref{app-fair08}, we also showed the fairness analysis of the hate detection algorithms using the $Pr(AAE \geq 0.8)$ threshold in order to provide equal comparisons and observed similar patterns.

A common first step in addressing biases in datasets is using a larger corpus to increase instances of the minority class \cite{park-etal-2018-reducing}. We evaluated the \textit{DWMW17} and \textit{FDCL18} datasets individually and created two aggregate datasets to better understand the racial biases and annotation agreement across the toxic language datasets. These experiments informed the challenges and feasibility of using a larger corpus or complimentary AAE toxic language data to address the racial biases present in available datasets. We chose four commonly used toxic language datasets that are taken from Twitter and annotated with compatible definitions of either \verb|toxic| or \verb|hate|. The low AAE totals in the individual datasets of \textit{Golbeck} and \textit{WH16} made it difficult to assess the racial biases on their own. When evaluated as part of the aggregated dataset, they informed cross-corpus label agreement and how the use of a larger corpus impacts classification and fairness metrics.

For evaluation, we segmented the datasets based on binary labels of \verb|hate| or \verb|toxic|. The \textit{Hate} dataset is aggregated on the \verb|hate| labels from \textit{DWMW17} and \textit{FDCL18} and  only the subset of positive instances of \textit{WH16} that were labeled as \verb|racist| or \verb|sexist|. The \textit{Toxic} aggregate dataset included the \verb|hate| and \verb|offensive| instances of \textit{DWMW17}, the \verb|hate| and \verb|abusive| instances from \textit{FDCL18}, the positive instances of \textit{WH16} \verb|racist| or \verb|sexist|, and the \verb|harassment| instances from \textit{Golbeck17}. We evaluated the outcomes from the hate detection algorithms applied to \textit{DWMW17} and \textit{FDCL18} datasets based on the \verb|toxic| label due to the strong association with AAE text being marked as toxic by annotators and the lower sample count of hate in AAE texts. Additionally, using the aggregate datasets of \textit{Hate} and \textit{Toxic}, we gain insight into the effects of using a larger corpus to reduce biases and compare the effects of annotation consistency on classifier performance. 

\begin{table*}[h]
  \begin{threeparttable}
  \caption{Datasets Characteristics and Dialect Label Comparisons}
  \label{tab:datasets}
  \begin{tabular}{lrrrrccc}
    \toprule
    Dataset (n=) & AAE\tnote{1} & Toxic & Hate & Total & $\text{Toxic}_{\text{AAE}}, \text{Toxic}_{\text{SAE}}$\tnote{2} & $\text{Hate}_{\text{AAE}},\text{Hate}_{\text{SAE}}$\tnote{3} & Annotated By\tnote{4} \\
\midrule
    \textit{DWMW17}				&	4878	& 20620 & 1430	& 22050 & 0.98, 0.80	& 0.04, 0.06 & 3+ CSW\\
    \textit{FDCL18}				&	1265	& 24821	& 4119	& 87371	& 0.84, 0.26	& 0.21, 0.04 & 5 CSW\\
    \textit{Golbeck} 			& 	92		& 4760	& -		& 19715 & 0.50, 0.24	& - & 2+ GRA\\
    \textit{WH16$_{subset}$} 	& 	12 		& 3360 	& 3360 	& 3360	& -				& - & Authors + DE \\
    \textit{Toxic} 				& 	6205	& 52679	& - 	& 134457& 0.94, 0.36 	& - & -\\
    \textit{Hate} 				&	6120 	& -		& 8710	& 123886& -				& 0.08, 0.07 & -\\
    \bottomrule    
  \end{tabular}
  {\footnotesize
  \begin{tablenotes}[flushleft,small]
    \item[1] Number of samples in the dataset Blodgett et al.'s \cite{blodgett-etal-2016-demographic} dialect model predicts as $Pr(AAE \geq 0.6)$
   \item [2] Proportion of the AAE vs. SAE dialect samples labeled as \verb|Toxic|
   \item [3] Proportion of the AAE vs. SAE dialect samples labeled as \verb|Hate|
   \item [4] Crowd-Sourced Worker (CSW), Graduate Research Assistant (GRA), Domain Expert (DE)
  \end{tablenotes}
  }
 \end{threeparttable}
\end{table*}

\section{Research Design}
\subsection{Models and Bias Remediation Techniques}
In prior work, a strong correlation was found between instances of AAE dialect and the associated annotation of toxic labels in abusive language datasets, which leads to racial bias against African American authors in models trained on them, introducing a higher false positive rate (FPR) for instances that are AAE \cite{sap-etal-2019-risk,davidson-etal-2019-racial}. A higher FPR means that non-toxic AAE texts are more likely to be misclassified as toxic by hate detection algorithms. This lexicographical and identity bias can then become embedded and further propagated through the classifiers that are trained with them.

In this paper, we benchmark classifier performance and fairness indicators across the datasets based on a series of different hate detection algorithms including baselines, logistic regression models with unigram and bigram encodings, TF-IDF, and GloVe embeddings \cite{pennington2014glove}. We also evaluated a vanilla-BERT classifier \cite{devlin-etal-2019-bert}, a commonly used language model in NLP classification tasks. We then evaluated the effectiveness of two in-process debiasing algorithms that use BERT as a base model. Finally, we introduce a new ensemble framework as a proposed method to remediate biases that may be attributed to low-resource contexts.

\subsubsection*{In-Process Debiasing Algorithms}
\label{debiasing-algorithms-desc}
\hfill\\
\textbf{Explanation Regularization:} Kennedy et al. \cite{kennedy-etal-2020-contextualizing} used the Occlusion (OC) and Sampling and Occlusion (SOC) explanation over BERT to generate hierarchical explanations for a prediction and use it to score how a phrase contributes to the classification. This score is then used to regularize the model during learning, with the intention of mitigating the compositional effects of a phrase and the context around it. AAE's characteristic of using the n-word to indicate another person is an example of the in-group reclaiming a group identifier that was used to oppress and dehumanize Black individuals. In-group usage of the term is not considered hateful, only out-group usage is deemed hateful. Ass Camouflage Construction (ACC) is another characteristic of AAE that has "ass" or "butt" usually proceeded by a possessive pronoun and is an equivalent to the reflexive self. With the purpose of reducing the classifier's bias towards the in-group identifiers and AAE pronoun altercations, we applied this regularization algorithm to help the model to learn the context surrounding the pronouns and identifiers in AAE \cite{green_2002,collins_aae,sweetland_aae}.

\noindent\textbf{MinDiff Framework:}
Prost et al. \cite{prost-2019} introduced a regularization technique that penalized models for dependence between the distribution of predicted probabilities and a protected subgroup, such as AAE. This framework attempts to minimize the difference between the protected subgroup and the majority group distributions. This algorithm minimizes the differences in FPR across the two slices with the intention of a minimal impact on classification performance. This algorithm has been discussed as an effective manner of reducing biases in language modeling tasks but is limited by the available data samples in the group slices.

\subsubsection*{Hierarchical Ensemble Framework (HxEnsemble)} 
\label{ensemble-desc}
\hfill\\
This paper proposes a hierarchical ensemble framework that minimizes potential classifier performance degradation while mitigating biases that are a result of training data that does not effectively represent the target population. To achieve closer to equality results across groups (AAE and SAE authors), we make use of the general classifier that contains biases to the "protected group", AAE authors. For instances predicted as positive (toxic) in the general model and in the dialect estimation model (as AAE), the ensemble will pass them to a specialized classifier that is pre-trained to the AAE dialect and fine-tuned on only AAE samples in the toxic language datasets. Effectively, in order to achieve a classifier that has closer to equal-outcomes across groups, we make use of an equity-based framework that is better able to predict positive instances of the protected group which the general model has been shown to exhibit bias against. 

A related technique was first presented by Howard et al. \cite{howard2017}, where it was shown that an ensemble framework achieved better classification performance and reduced FPR for misclassified emotions by using a combination of a generalized model and a specialized learner that is trained on the classes that are most commonly misclassified. As demonstrated in Sap et al. \cite{sap-etal-2019-risk}, annotator biases towards the AAE dialect may be minimized during labeling when the annotators are racially primed of the potential race of the author. Since AAE has different syntactical and lexicographical characteristics, we hypothesize that similar to racially primed annotators, a classifier that is pre-trained on non-toxic AAE samples is better able to distinguish between AAE samples in the toxic language datasets that are true-positives and those that are true-negatives. While the underlying language model remains a black-box, we conjecture that the AAE-BERT will have a better understanding of the AAE dialect. Therefore, when the classifier for AAE toxic language classification is fine-tuned on it, there will be less biases propagated than using a base language model that is trained on a corpus of mainly SAE samples.

\begin{figure}[h]
  \centering
  \includegraphics[width=\linewidth]{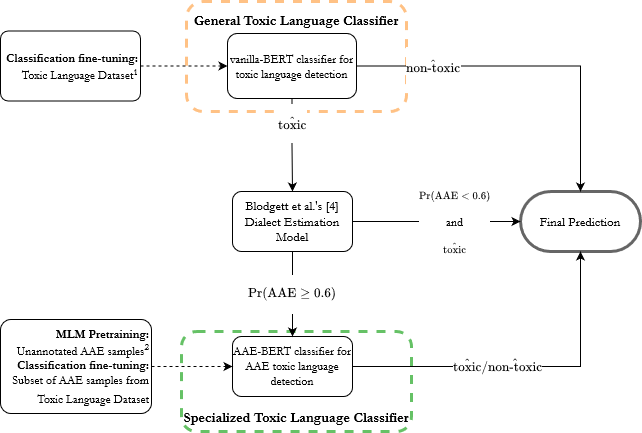}
  \caption{
  The Hierarchical Ensemble (HxEnsemble) Framework \\
  $^1$ Training set of DWMW17, FDCL18, Toxic, or Hate is used. More details about these datasets can be found in Table \ref{tab:datasets}. \\
  $^2$ The unannotated AAE samples used for the MLM pre-training come from Blodgett et al. \cite{blodgett-etal-2016-demographic}
  }
  \label{ensemble-fig}
\end{figure}

The proposed HxEnsemble (Figure \ref{ensemble-fig}) uses a general toxic language detection model, in our case a vanilla-BERT classifier trained on the original dataset. If the general model predicted a positive instance (\verb|hate| or \verb|toxic|), we then used the out-of-box Blodgett et al. dialect estimation model \cite{blodgett-etal-2016-demographic} to predict the probability the sample is AAE. If the sample is not AAE, we returned the predicted result from the general classifier. However, if the text is AAE we then passed it through the specialized AAE classifier and the ensemble model returned that prediction. The specialized classifier is created by fine-tuning the vanilla BERT on a masked language model task with Blodgett et al.'s corpus of demographic based Twitter data that belongs to African American authors \cite{blodgett-etal-2016-demographic}. We then use this BERT-AAE language model to  fine-tune a classifier head on the AAE samples in the same train-validation-test splits that the general model was trained on. The final prediction for the positive AAE sample is the prediction of the specialized AAE classifier. 

\subsection{Fairness Metrics}
\label{fairness-metrics-descr}
Fairness metrics are used to statistically evaluate notions of fairness in classifier performance, where certain metrics can reflect different definitions of fairness \cite{verma_rubin_2018}. In previous work by Zhou et al. \cite{zhou-etal-2021-challenges} and Xia et al. \cite{xia-etal-2020-demoting} that explore bias remediation techniques for toxic language detection, the FPR is used as the single fairness metric for reporting how the biases in the training data propagate to the models. As seen in Table \ref{tab:datasets}, the positively-skewed distribution of AAE texts in the datasets means that there is a very limited number of true-negatives in the data splits that were evaluated and the insights to the effects of biases is limited.  As such and in order to address the low occurrences of AAE samples, we also compute a fairness metrics based on the disparate impact (DI) metric \cite{verma_rubin_2018}, also commonly referred to as adverse impact. DI is a fairness metric that evaluates the predictive parity ratio to compare predicted outcomes across groups, which does not rely on the annotations in its computation. In practice, the acceptable fairness range for this metric is limited to $\{0.8 - 1.2\}$ \cite{adverseimpact}, and a $DI < 1$ is interpreted as bias against the protected group, in our case the AAE authors. Conversely, if $DI > 1$ there is said to be bias towards the protected group.

\begin{equation}
DI_{\text{fav}} = \frac{Pr(\hat{Y} = 0 | D = \text{AAE})}{Pr(\hat{Y} = 0 | D = \text{SAE})}, DI_{\text{unfav}} = \frac{Pr(\hat{Y} = 1 | D = \text{AAE})}{Pr(\hat{Y} = 1 | D = \text{SAE})}
\end{equation}

In toxic language detection, the notion of a favorable prediction is subjective. Therefore we analyzed the DI for when the prediction is non-toxic ($\text{DI}_{\text{fav}}$) and toxic ($\text{DI}_{\text{unfav}}$) to gain more insight into how the classifier treats the different dialect groups for both outcomes. Lastly, we looked at false negative rates (FNR) to provide insight into whether the bias remediation techniques are causing disagreement with the positively-skewed toxic annotations for AAE authors. 

\subsection{Experiment Implementation}
To evaluate the hate detection algorithms as applied to the aforementioned datasets, we randomly split all our datasets stratified on the positive-cases into 80\% training, 10\% validation, and 10\%  for testing. We used the training and validation splits to run a grid-search in order to fine-tune the algorithms we benchmarked. We chose the hyperparameters that resulted in the best validation F1-score and trained the model across 10 random seeds to report our results below (\S\ref{results-analysis}).\footnote{All our code is available at \url{https://github.com/matanhalevy/DebiasingHateDetectionAAE}}

For the logistic regression-based models, we tune the hyperparameters on learning rates $\{2\mathrm{e}{-3}, 2\mathrm{e}{-5}, 5\mathrm{e}{-5}\}$, epochs $\{10, 100, 1000\}$, and batch sizes $\{16,32,64\}$. For all BERT-based models, we tuned the hyperparameters on learning rates $\{2\mathrm{e}{-3}, 2\mathrm{e}{-5}, 5\mathrm{e}{-5}\}$, epochs $\{1, 2, 3\}$, and batch sizes $\{16,32,64\}$. For the Explanation Regularization model, we searched the regularization strength $\{0.1, 0.3, 0.5\}$, and for the MinDiff Framework we fine-tuned MinDiff weight to $\{0.5, 1, 1.5\}$. With the specialized AAE learner in the hierarchical ensemble, we ran the parameter search on both the masked language model task and classification task together.

\section{Results and Analysis}
\label{results-analysis}
All metrics reported below include the subscript of their standard deviation across the 10 randomized trials.

\subsection{DWMW17}

In Table \ref{tab:dwmw17}, we present the results of the different classifiers trained on \textit{DWMW17}. Individually, this dataset is the most difficult to analyze for fairness due to the fact that 98\% of AAE samples in the complete dataset are annotated as toxic, and only $2.5\%$ of the 502 AAE samples in the test set are labeled as true negatives. The classifiers of BERT, BERT with OC and SOC, and our HxEnsemble method have the best classification performance. After evaluating the fairness metrics across the best performing models, we verify that the HxEnsemble model achieved the best disparate impact scores for prediction of non-toxic and toxic outcomes. We theorize this result is due to the slight increase in the $\text{FNR}_{AAE}$ and decrease in the $\text{FNR}_{SAE}$ compared to the other models. We observe that BERT+OC achieves the lowest $\text{FPR}_{AAE}$. However, for BERT+OC, BERT+SOC, and HxEnsemble, the $\text{FNR}_{AAE}$ and $\text{FPR}_{AAE}$ are within each others' error bounds. Overall we note that for \textit{DWMW17}, all classifiers are biased towards being more likely to predict AAE text as toxic by a significant ratio, which is not as evident when examining the prediction parity across groups. Due to the low true-negative AAE samples, we did not conclude any significant results for the classifiers trained on \textit{DWMW17} with regards to unfavorable outcomes. 

\begin{table*}[h]
{\small
\begin{threeparttable}
\caption{DWMW17 Results}
\label{tab:dwmw17}
\begin{tabular}{@{}lrrrrrrrr@{}}
\toprule
Task Name \tnote{*}     & Acc                 & F1                  & $\text{DI}_{\text{fav}}$ \tnote{1}         & $\text{DI}_{\text{unfav}}$\tnote{2}        & $\text{FNR}_{\text{AAE}}$\tnote{3}         & $\text{FNR}_{\text{SAE}}$\tnote{4}         & $\text{FPR}_{\text{AAE}}$\tnote{5}         & $\text{FPR}_{\text{SAE}}$ \\ \midrule
N-Gram         & $0.947_{\pm 0.001}$ & $0.968_{\pm 0.000}$   & $0.112_{\pm 0.006}$ & $1.203_{\pm 0.003}$ & $0.003_{\pm 0.001}$ & $0.027_{\pm 0.001}$ & $0.308_{\pm 0.000}$ & $0.206_{\pm 0.004}$ \\
TF-IDF         & $0.872_{\pm 0.000}$ & $0.926_{\pm 0.000}$   & $0.100_{\pm 0.009}$ & $1.111_{\pm 0.001}$ & $0.007_{\pm 0.001}$ & $0.035_{\pm 0.001}$ & $0.846_{\pm 0.000}$ & $0.604_{\pm 0.003}$ \\
GloVe          & $0.886_{\pm 0.002}$ & $0.933_{\pm 0.001}$   & $0.137_{\pm 0.013}$ & $1.177_{\pm 0.021}$ & $0.011_{\pm 0.003}$ & $0.062_{\pm 0.010}$ & $0.523_{\pm 0.061}$ & $0.421_{\pm 0.046}$ \\
\textit{BERT}           & $0.965_{\pm 0.002}$ & $0.979_{\pm   0.001}$ & $0.098_{\pm 0.014}$ & $1.231_{\pm 0.006}$ & $0.002_{\pm 0.001}$ & $0.024_{\pm 0.003}$ & $0.308_{\pm 0.115}$ & $0.108_{\pm 0.012}$ \\
\textit{BERT+OC}      & $0.966_{\pm 0.001}$ & $0.979_{\pm 0.001}$   & $0.100_{\pm 0.005}$ & $1.248_{\pm 0.008}$ & $0.002_{\pm 0.001}$ & $0.030_{\pm 0.003}$ & $\mathbf{0.238_{\pm 0.044}}$ & $\mathbf{0.081_{\pm 0.013}}$ \\
\textit{BERT+SOC}     & $0.968_{\pm 0.001}$ & $0.980_{\pm 0.001}$   & $0.098_{\pm 0.008}$ & $1.248_{\pm 0.005}$ & $0.002_{\pm 0.001}$ & $0.029_{\pm 0.002}$ & $0.246_{\pm 0.049}$ & $0.077_{\pm 0.009}$ \\
BERT+MD & $0.887_{\pm 0.059}$ & $0.936_{\pm   0.031}$ & $0.053_{\pm 0.057}$ & $1.108_{\pm 0.119}$ & $0.003_{\pm 0.007}$ & $0.016_{\pm 0.026}$ & $0.746_{\pm 0.310}$ & $0.601_{\pm 0.420}$ \\
\textit{HxEnsemble}     &  $0.964_{\pm 0.002}$ &$0.978_{\pm 0.001}$ &$\mathbf{0.114_{\pm 0.008}}$ & $\mathbf{1.223_{\pm 0.004}}$ & $\mathbf{0.004_{\pm 0.002}}$ & $\mathbf{0.022_{\pm 0.001}}$ & $0.262_{\pm 0.040}$  & $0.121_{\pm 0.009}$  \\ \bottomrule
\end{tabular}
{\footnotesize
\begin{tablenotes}[flushleft]
    \item[*] Models that had the best classification accuracy are \textit{italicized} and the best fairness indicators on the best performing models are \textbf{bolded} per column.
    \item[1] Disparate Impact for favorable outcomes measures the prediction disparity for AAE and SAE authors being predicted as non-toxic.
    \item[2] Disparate Impact for unfavorable outcomes measures the prediction disparity for AAE and SAE authors being predicted as toxic.
    \item[3] $\text{FNR}_{\text{AAE}}$: the "best" fairness metric in this case is the highest FNR, since the annotations are biased to labeling AAE samples as toxic, an increase may be indicative of the model unlearning these biases.
    \item[4] $\text{FNR}_{\text{SAE}}$: for this metric, we want the lowest score as we only care about classification performance of the SAE group.
    \item[5] For both AAE and SAE group, a lower FPR is better.
\end{tablenotes}
}
 \end{threeparttable}
 }
\end{table*}

\subsection{FDCL18}
Table \ref{tab:fdcl18} shows the results the classifiers achieved on \textit{FDCL18}. Similar to \textit{DWMW17}, there is a low true-negative count in the test set with $14.5\%$ of 117 AAE samples being annotated as non-toxic. BERT, BERT+OC, BERT+SOC, and HxEnsemble achieved the best classifier performance. HxEnsemble achieved the best fairness results across favorable and unfavorable disparate impact scores and $\text{FPR}_{\text{AAE}}$. It's worth noting that HxEnsemble also has the highest $\text{FNR}_{AAE}$ amongst this subset and the lowest disparity between $\text{FPR}_{AAE}$ to $\text{FPR}_{SAE}$, providing some empirical evidence that increasing the language model's concept of the AAE dialect causes the model to disagree with the biased AAE annotations. 

We noted that for all models the $\text{FNR}_{\text{AAE}}$ is lower than $\text{FNR}_{\text{SAE}}$, while the $\text{FPR}_{\text{AAE}}$ is larger than $\text{FPR}_{\text{SAE}}$. This means that all models are more likely to misclassify non-toxic AAE samples as toxic compared to SAE samples and less likely to misclassify toxic AAE samples as non-toxic compared to SAE samples. The high $\text{FPR}_{\text{AAE}}$ and disparate impact scores show significant bias towards AAE authors for both favorable and unfavorable outcomes across all models that are trained on FDCL18. For this dataset, the bias remediation techniques helped reduce the FPR disparity but did not effectively mitigate the prediction disparity. These results demonstrated how underlying data bias to AAE authors propagate to the models even with bias-remediation techniques.

\begin{table*}[h]
{\small
\begin{threeparttable}
\caption{FDCL18 Results}
\label{tab:fdcl18}
\begin{tabular}{@{}lrrrrrrrr@{}}
\toprule
Task Name \tnote{*}     & Acc                 & F1                  & $\text{DI}_{\text{fav}}$        & $\text{DI}_{\text{unfav}}$        & $\text{FNR}_{\text{AAE}}$        & $\text{FNR}_{\text{SAE}}$        & $\text{FPR}_{\text{AAE}}$        & $\text{FPR}_{\text{SAE}}$ \\ \midrule
N-Gram         & $0.935_{\pm 0.000}$ & $0.878_{\pm 0.001}$   & $0.194_{\pm 0.000}$ & $3.398_{\pm 0.016}$ & $0.040_{\pm 0.000}$ & $0.147_{\pm 0.002}$ & $0.235_{\pm 0.000}$ & $0.036_{\pm 0.001}$  \\
TF-IDF         & $0.896_{\pm 0.001}$ & $0.788_{\pm 0.001}$   & $0.334_{\pm 0.005}$ & $3.453_{\pm 0.021}$ & $0.187_{\pm 0.005}$ & $0.291_{\pm 0.002}$ & $0.294_{\pm 0.000}$ & $0.036_{\pm 0.000}$  \\
GloVe          & $0.900_{\pm 0.001}$ & $0.810_{\pm 0.002}$   & $0.209_{\pm 0.005}$ & $3.417_{\pm 0.028}$ & $0.064_{\pm 0.005}$ & $0.222_{\pm 0.005}$ & $0.294_{\pm 0.000}$ & $0.056_{\pm 0.002}$  \\
\textit{BERT}           & $0.943_{\pm 0.001}$ & $0.896_{\pm   0.001}$ & $0.155_{\pm 0.005}$ & $3.287_{\pm 0.015}$ & $0.001_{\pm 0.003}$ & $\mathbf{0.098_{\pm 0.003}}$ & $0.229_{\pm 0.019}$ & $0.043_{\pm 0.001}$ \\
\textit{BERT+OC}      & $0.943_{\pm 0.001}$ & $0.895_{\pm 0.001}$   & $0.164_{\pm 0.009}$ & $3.325_{\pm 0.038}$ & $0.006_{\pm 0.007}$ & $0.108_{\pm 0.003}$ & $0.206_{\pm 0.031}$ & $\mathbf{0.040_{\pm 0.002}}$ \\
\textit{BERT+SOC}     & $0.943_{\pm 0.001}$ & $0.895_{\pm 0.002}$   & $0.166_{\pm 0.016}$ & $3.315_{\pm 0.038}$ & $0.009_{\pm 0.009}$ & $0.107_{\pm 0.005}$ & $0.212_{\pm 0.041}$ & $0.040_{\pm 0.003}$  \\
BERT+MD & $0.824_{\pm 0.102}$ & $0.429_{\pm   0.453}$ & $0.616_{\pm 0.406}$ & $1.465_{\pm 1.551}$ & $0.526_{\pm 0.500}$ & $0.555_{\pm 0.469}$ & $0.082_{\pm 0.101}$ & $0.035_{\pm 0.045}$ \\
\textit{HxEnsemble}     &  $0.943_{\pm 0.000}$ & $0.896_{\pm 0.001}$ &  $\mathbf{0.180_{\pm 0.019}}$ & $\mathbf{3.243_{\pm 0.051}}$  & $\mathbf{0.016_{\pm 0.015}}$ &                  $0.101_{\pm 0.004}$ & $\mathbf{0.183_{\pm 0.035}}$ & $0.045_{\pm 0.009}$ \\ \bottomrule
\end{tabular}
{\footnotesize
\begin{tablenotes}[flushleft, online]
    \item[*] Please refer to Table \ref{tab:dwmw17} for the explanation of the table results.
\end{tablenotes}
}
 \end{threeparttable}
 }
\end{table*}

\subsection{Toxic}

\begin{table*}[h]
{\small
\begin{threeparttable}
\caption{Toxic Results}
\label{tab:toxic}
\begin{tabular}{@{}lrrrrrrrr@{}}
\toprule
Task Name \tnote{*}     & Acc                 & F1                  & $\text{DI}_{\text{fav}}$        & $\text{DI}_{\text{unfav}}$        & $\text{FNR}_{\text{AAE}}$        & $\text{FNR}_{\text{SAE}}$        & $\text{FPR}_{\text{AAE}}$        & $\text{FPR}_{\text{SAE}}$ \\ \midrule
N-Gram         & $0.900_{\pm 0.000}$ & $0.867_{\pm 0.001}$   & $0.059_{\pm 0.001}$ & $2.904_{\pm 0.015}$ & $0.014_{\pm 0.001}$ & $0.189_{\pm 0.003}$ & $0.385_{\pm 0.000}$ & $0.054_{\pm 0.001}$ \\
TF-IDF         & $0.844_{\pm 0.001}$ & $0.780_{\pm 0.002}$   & $0.136_{\pm 0.002}$ & $3.121_{\pm 0.026}$ & $0.077_{\pm 0.001}$ & $0.321_{\pm 0.004}$ & $0.473_{\pm 0.019}$ & $0.065_{\pm 0.002}$ \\
GloVe          & $0.840_{\pm 0.000}$ & $0.787_{\pm 0.000}$   & $0.114_{\pm 0.003}$ & $2.752_{\pm 0.019}$ & $0.061_{\pm 0.002}$ & $0.264_{\pm 0.002}$ & $0.615_{\pm 0.000}$ & $0.106_{\pm 0.002}$  \\
\textit{BERT}           & $0.915_{\pm 0.000}$ & $0.890_{\pm 0.001}$   & $0.058_{\pm 0.003}$ & $2.850_{\pm 0.012}$ & $0.008_{\pm 0.001}$ & $0.156_{\pm 0.002}$ & $0.300_{\pm 0.040}$ & $\mathbf{0.046_{\pm 0.001}}$  \\
BERT+OC      & $0.911_{\pm 0.001}$ & $0.883_{\pm 0.001}$   & $0.053_{\pm 0.004}$ & $2.863_{\pm 0.031}$ & $0.009_{\pm 0.001}$ & $0.165_{\pm 0.006}$ & $0.392_{\pm 0.065}$ & $0.050_{\pm 0.003}$  \\
BERT+SOC     & $0.911_{\pm 0.001}$ & $0.883_{\pm 0.001}$   & $0.059_{\pm 0.003}$ & $2.837_{\pm 0.025}$ & $0.010_{\pm 0.001}$ & $0.162_{\pm 0.005}$ & $0.319_{\pm 0.036}$ & $0.051_{\pm 0.003}$  \\
BERT+MD & $0.632_{\pm 0.131}$ & $0.227_{\pm   0.294}$ & $0.700_{\pm 0.483}$ & $0.410_{\pm 0.531}$ & $0.637_{\pm 0.480}$ & $0.629_{\pm 0.486}$ & $0.362_{\pm 0.480}$ & $0.343_{\pm 0.457}$  \\
\textit{HxEnsemble}     &  $0.914_{\pm 0.001}$	& $0.887_{\pm 0.001}$ & 	$\mathbf{0.072_{\pm 0.005}}$ &	$\mathbf{2.778_{\pm 0.029}}$	& $\mathbf{0.016_{\pm 0.003}}$ &	$\mathbf{0.152_{\pm 0.004}}$ & $\mathbf{0.277_{\pm 0.016}}$	& $0.052_{\pm 0.003}$ \\ \bottomrule
\end{tabular}
{\footnotesize
\begin{tablenotes}[flushleft, online]
    \item[*] Please refer to Table \ref{tab:dwmw17} for the explanation of the table results.
\end{tablenotes}
}
 \end{threeparttable}
}
\end{table*}

In Table \ref{tab:toxic}, the results for the classifiers on the aggregate \textit{Toxic} dataset is shown, we note that for this aggregate dataset the true-negative count for AAE instances is $5.7\%$. BERT and HxEnsemble are the algorithms that had the best classification scores, while BERT with OC and SOC achieved slightly worse results. HxEnsemble achieved the lowest $\text{FPR}_{\text{AAE}}$ across all algorithms, and the best disparate impact scores across all BERT-based models. MinDiff performs poorly on this dataset, even with more AAE samples available. This pattern is demonstrated across the low F1 scores MinDiff has on every dataset, excluding \textit{DWMW17}. The same pattern of higher FPR and lower FNR for AAE samples exists in this aggregate but, compared to the standalone \textit{FDCL18}, the disparate impact scores provide partial evidence that an effective strategy to mitigate biases is to aggregate a larger corpus. In comparison to Table \ref{tab:hate}, the F1 scores for the algorithms suggest that there is more label agreement for the definition of \verb|toxic|, compared to the F1 scores on the more stringent definition of \verb|hate|. This is indicative that in order to better deal with label inconsistency issues across toxic-language datasets, users must opt for a more general definition of toxic rather than hate.

\subsection{Hate}

As in the \textit{Toxic} dataset, BERT and HxEnsemble perform best on the \textit{Hate} aggregate dataset. However, across all algorithms the low F1-score suggested the aggregated datasets individually had a low agreement in their \verb|hate| label. Although we attribute this to poor model performance, the favorable and unfavorable disparate impact scores achieved more fairness in their predictions than in other datasets. HxEnsemble introduces bias in favor of AAE as the high FNR and low FPR yielded a lower disparate impact score for the unfavorable outcome compared to vanilla-BERT. The use of this aggregate dataset is not very informative of biases against AAE authors as the learners themselves struggle to deal with the class-imbalance and inconsistent hate definitions across the datasets as seen by low classification scores.

\begin{table*}[h]
{\small
\begin{threeparttable}
\caption{Hate Results}
\label{tab:hate}
\begin{tabular}{@{}lrrrrrrrr@{}}
\toprule
Task Name \tnote{*}     & Acc                 & F1                  & $\text{DI}_{\text{fav}}$        & $\text{DI}_{\text{unfav}}$        & $\text{FNR}_{\text{AAE}}$        & $\text{FNR}_{\text{SAE}}$        & $\text{FPR}_{\text{AAE}}$        & $\text{FPR}_{\text{SAE}}$ \\ \midrule
N-Gram         & $0.936_{\pm 0.000}$ & $0.368_{\pm 0.006}$   & $0.990_{\pm 0.002}$ & $1.316_{\pm 0.074}$ & $0.736_{\pm 0.026}$ & $0.735_{\pm 0.008}$ & $0.022_{\pm 0.001}$ & $0.013_{\pm 0.001}$ \\
TF-IDF         & $0.936_{\pm 0.000}$ & $0.234_{\pm 0.002}$   & $0.985_{\pm 0.001}$ & $2.156_{\pm 0.071}$ & $0.807_{\pm 0.008}$ & $0.863_{\pm 0.002}$ & $0.015_{\pm 0.000}$ & $0.004_{\pm 0.000}$ \\
GloVe          & $0.928_{\pm 0.003}$ & $0.181_{\pm 0.034}$   & $1.001_{\pm 0.006}$ & $0.947_{\pm 0.381}$ & $0.959_{\pm 0.033}$ & $0.881_{\pm 0.031}$ & $0.015_{\pm 0.006}$ & $0.011_{\pm 0.005}$ \\
\textit{BERT}           & $0.943_{\pm 0.001}$ & $\mathbf{0.522_{\pm 0.006}}$   & $\mathbf{0.987_{\pm 0.005}}$ & $\mathbf{1.231_{\pm 0.084}}$ & $0.524_{\pm 0.027}$ & $\mathbf{0.534_{\pm 0.008}}$ & $0.035_{\pm 0.004}$ & $0.024_{\pm 0.002}$ \\
BERT+OC      & $0.938_{\pm 0.002}$ & $0.386_{\pm 0.031}$   & $0.979_{\pm 0.016}$ & $1.675_{\pm 0.499}$ & $0.690_{\pm 0.090}$ & $0.724_{\pm 0.035}$ & $0.029_{\pm 0.013}$ & $0.011_{\pm 0.004}$ \\
BERT+SOC     & $0.937_{\pm 0.002}$ & $0.396_{\pm 0.019}$   & $0.971_{\pm 0.029}$ & $1.890_{\pm 0.954}$ & $0.645_{\pm 0.111}$ & $0.710_{\pm 0.028}$ & $0.037_{\pm 0.023}$ & $0.013_{\pm 0.003}$ \\
BERT+MD & $0.930_{\pm 0.000}$ & $0.079_{\pm   0.068}$ & $0.400_{\pm 0.516}$ & $0.600_{\pm 0.516}$ & $0.400_{\pm 0.516}$ & $0.400_{\pm 0.516}$ & $0.600_{\pm 0.516}$ & $0.600_{\pm 0.516}$  \\
\textit{HxEnsemble}     &  $0.942_{\pm 0.001}$ &	$0.516_{\pm 0.009}$ &	$1.030_{\pm 0.011}$	& $0.445_{\pm 0.203}$&$\mathbf{0.795_{\pm 0.101}}$	&$0.549_{\pm 0.011}$&	$\mathbf{0.008_{\pm 0.004}}$	&$\mathbf{0.021_{\pm 0.001}}$\\ \bottomrule
\end{tabular}
{\footnotesize
\begin{tablenotes}[flushleft, online]
    \item[*] Please refer to Table \ref{tab:dwmw17} for the explanation of the table results.
\end{tablenotes}
}
 \end{threeparttable}
}
\end{table*}

\section{Error Analysis and Challenges for Hate Detection Algorithms in Classifying AAE Text}
\label{error-analysis}

\begin{table*}[h]
\caption{AAE Tweets that were Misclassified by HxEnsemble} 
\label{tab:misclassified}
\begin{tabular}{@{}lp{120mm}|cc@{}}
\toprule
 & Tweet &  $\hat{y}$ &  $y$ \\
\midrule
1 & RT : still stuntin on my ex bitch &     1 &      0 \\
2 & RT : IF U SEE ME SMILING IM EITHER ABOUT TO KILL U OR REALLY FUCKED UP EITHER WAY I AINT IN THE RITE STATE OF MIND &     0 &      1 \\
3 & RT : That ain't yo n**** if he let you go to sleep mad sis &     1 &      0 \\
4 & ID SLAP YO BIPLAR ASS POLAR BITCH U AINT SHIT BUT A WANNA BE ASS BITCH WHO BETTER WATCH THE FUCK OUT YOU UGLY ASS HOE &     1 &      0 \\
5 & RT : Whatchu know bout livin in the ghetto, eating cottage cheese 3 times a day. Lil salt Lil peppa lil A1, Lil barbeeequeee &     0 &      1 \\
6 & Yea all these bad ass bey bey kids ride bikes over here &     1 &      0 \\
7 & Y'all n****s annoying I'm bout to start calling y'all sis &     1 &      0 \\
8 & Blatt trash too ... but twitter said he was good &     0 &      1 \\
9 & YO IMMA MOTHERFUCKING BOYYYYYYYYYYYYYYYYY &     0 &      1 \\
10 & Fucc u primo aint nobody ask u 2 fav dink lies &     0 &      1 \\
11 & RT : Who did y'all n****s give y'all hearts to for y'all to be so cold hearted &     1 &      0 \\
12 & You gots da BET and da MSNBC wit dat colored reverend who only listen to coloreds. What else I missing? &     0 &      1 \\
13 & He did go in i was actually surprised he when that hard RT : N the crib listen 2 my n** G-Eazy;don't diss my homie &     0 &      1 \\
14 & Dumb Haitian fake black faggots. Go to Haiti and neck yourself. &     0 &      1 \\
15 & \#TripleTalaqArrest 2fgt dis social menace \#TripleTalaq rgts shd b gvn 2d Muslim wmn2, den dis evil wl hv a natural death \#TripleTalaqDebate &     0 &      1 \\
16 & yall trash but  cold &     0 &      1 \\
17 & RT : Whatchu sayin hoe, YOU KNOW IM THE MAN HOE &     1 &      0 \\
\bottomrule
\end{tabular}
\end{table*}

In Table \ref{tab:misclassified} we present all the misclassified AAE tweets from our test set from a run using our HxEnsemble. Below, we categorized these tweets into 7 categories in order to better understand the challenges in debiasing hate-detection classifiers as applied to AAE authors. We include the False-Negative (FN) and False-Positive (FP) prepended to our category names to clarify the type of challenge.

\begin{enumerate}
\item\textbf{FN: Mislabeled, Non-Toxic} HxEnsemble predicted examples 5, 9, 13, and 16 as non-toxic while annotators labeled them as \verb|toxic|. For these examples we see that the HxEnsemble model overcame the annotators' bias and provided the prediction that matches the definition of non-toxic language provided to the annotators.

\item\textbf{FN: Mislabeled, Non-Targeted Threat} For example 2, while the text contains aggressive language it remains non-targeted and the author of the tweet is only talking about themselves. We disagreed with the annotators that label this  as \verb|toxic| and agree with the HxEnsemble prediction. However, this may be interpreted as a threat but, since it's not directed at a person or group, we dismiss it as such. 

\item\textbf{FN: Missing Context, Unclear Toxicity} We observed that instances 8 and 10 are not clear in what constituted them as \verb|toxic|. While these two examples have sentiments that may be slightly demeaning, we were unable to conclude whether we agreed with the annotators' labels. 

\item\textbf{FN: Toxic and False-Positive AAE} Three examples of \verb|toxic| instances were misclassified as AAE but the dialect model used the lower threshold of $0.6$. However, had we used the threshold of $0.8$ these would not have been classified as AAE. For 12 and 14, the general classifier predicted them as \verb|toxic| and the specialized classifier predicted them as \verb|non-toxic|. We hypothesize this is because these two examples are \verb|toxic| to Black people, and the specialized classifier did not have many examples of that type of toxicity in the AAE training dataset. For instance 15, lexical variation may have been used as a common method to avoid moderation detection \cite{chancellor_2016} as \verb|toxic| by the general classifier.

\item\textbf{FP: Mislabeled, Toxic Tweets} Our team found only one false-positive example (instance 4) that was predicted by HxEnsemble to be \verb|toxic| when the annotators did not label it as such. This tweet is directed at someone and harasses them based on a mental-health condition and calls them several derogatory curse words. We found the model predicted this example correctly while annotation across all datasets included in the aggregate should have labeled it as \verb|toxic| as well. 

\item\textbf{FP: Use of Pejorative for Third-Parties} For instances 1, 3, 7, and 17, our team found that these four examples the authors all referred to a third-party by either using the n-word or using the b-word to speak negatively about them. Instance 1 in particular is a retweeted tweet that is very similar to a popular rap lyric, "I'm just stuntin' on my ex-bitch." \cite{21savage} This suggests that even in cases that may not be toxic, the model struggles to differentiate between toxicity and use of these words in negative sentiment when used to describe a third-party. 

\item\textbf{FP: Use of Curse Words in Neutral Context} In examples 6 and 11, the false positives are resulted from the HxEnsemble unable to understand the Ass Camouflage Construction (ACC) and the n-word to replace an equivalent "guys" commonly used in AAE without implying toxic connotation. 

\end{enumerate}

As seen in the above challenges and summarized in Table \ref{tab:misclassified-breakdown}, we attributed $35.3\%$ of the AAE misclassified instances by HxEnsemble to incorrect annotations in the datasets. We also observed that 11.8\% of the misclassified AAE samples were due to non-AAE samples being misclassified by the dialect model, when the general classifier correctly predicted the texts as toxic. While the subjectivity of some challenge types may mean that there is higher disagreement with the labels provided in the datasets. These findings demonstrated how the efficacy of bias-mitigation strategies in addressing annotation biases may be under-reported using standard classification and fairness metrics. We also note that while the HxEnsemble method was able to effectively mitigate some of the biases towards AAE, it still struggled at times with some AAE characteristics such as ACC and the usage of the n-word. Lastly, we note that effects of incorrect annotations of AAE texts on the performance of the specialized AAE classifier and its ability to remediate the biases more effectively. 

\begin{table*}[h]
\caption{Breakdown of challenges in AAE Hate-Detection} 
\label{tab:misclassified-breakdown}
\begin{tabular}{@{}lrr@{}}
\toprule
Challenge Type                          & (n = ) & \% \\ \midrule
FN: Mislabeled, Non-Toxic              & 4                 & 23.5                   \\
FP: Use of Pejorative for Third-Parties & 4                 & 23.5                   \\
FN: Toxic and False-Positive AAE        & 3                 & 17.6                   \\
FP: Curse Words in Neutral Contexts     & 2                 & 11.8                   \\
FN: Missing Context, Unclear Toxicity   & 2                 & 11.8                   \\
FN: Mislabeled, Non-Targeted Threat        & 1                 & 5.9                    \\
FP: Mislabeled, Toxic                  & 1                 & 5.9                    \\ \bottomrule
\end{tabular}
\end{table*}

\section{Discussion}
In our investigation, we evaluated various hate detection algorithms with a focus on examining the effectiveness of our proposed hierarchical ensemble model and two bias-mitigation algorithms that addressed the racial biases present in toxic language datasets. In our experiments,  we observed that our hierarchical ensemble model consistently achieved state-of-the-art classification results while improving upon the fairness metrics we evaluated on. The explanation regularization technique was also able to reduce the biases against AAE authors better than the BERT-vanilla model without a dramatic decrease in classification performance. Across all datasets, the MinDiff regularization framework consistently performed worse than other classifiers, and while it had a closer false-positive error rate balance, this came at the cost of classification accuracy to the SAE authored tweets, rather than an improvement to AAE authored tweets. Since our HxEnsemble proposed framework is model agnostic, combining it with other bias-mitigation techniques to better address the complex underlying reasons for model biases can be investigated in future work.

Table \ref{tab:misclassified-breakdown} suggests that the hierarchical ensemble model increased FNR is strongly correlated with annotations that we deemed as misclassified as \verb|toxic| by annotators. This further strengthens the importance of utilizing additional metrics when evaluating debiasing methods. In fact, when investigating the effects of annotation bias, understanding the prediction disparities and classification inaccuracies provides valuable insight into the fairness of the underlying black-box models. We noted that at minimum, $35.3\%$ of misclassified AAE tweets by our HxEnsemble are mislabeled by original annotators. As a result, the effectiveness of the bias-mitigation strategies is likely a lower-bound estimate and provided empirical evidence that the increased FNR of our hierarchical ensemble model successfully mitigates some of the annotation biases towards AAE authors in toxic-language datasets.

For \textit{DWMW17}, \textit{FDCL18}, and \textit{Toxic} datasets we observed how disparate impact of non-toxic predictions strongly favors SAE, while for toxic prediction AAE are more likely than SAE. Additionally, almost all classifiers have $\text{FNR}_\text{SAE}$ computed larger than $\text{FNR}_\text{AAE}$, while the $\text{FPR}_\text{SAE}$ is smaller than $\text{FPR}_\text{AAE}$. While the biases of the dataset are present and perpetuated, HxEnsemble and the explanation regularization methods were able to decrease the disparities in these fairness metrics with minimal impact to classification metrics. This shows that on their own, bias-mitigation strategies are not enough to correct the underlying biased data and to fully address this issue in future work, efforts to relabel the dataset or create a new dataset that is less biased to AAE authors is essential.

We saw that using the aggregated dataset \textit{Toxic} provided mixed results for addressing biases. A decrease in the unfavorable disparate impact metric across all the classifiers is observed in the \textit{Toxic} dataset compared to the \textit{FDCL18}, the largest dataset in the aggregate. However, a slight increase in the $\text{FPR}_\text{AAE}$ and a decrease in the disparate impact of a favorable outcome indicates that using a larger corpus is an insufficient bias mitigation strategy for this issue. Additionally, label inconsistencies in the \verb|hate| annotation resulted in classifiers that had very poor performance and struggled to overcome the class imbalance in \textit{Hate}. Comparatively, using a less stringent definition of \verb|toxic| in the \textit{Toxic} aggregate allowed the classifiers to achieve strong performance. If a supplementary dataset of AAE authors for hate detection is created to address the racial biases present, ensuring that annotation consistency with other datasets will be a challenging issue to address. 

The biases against AAE dialect may be correlated to other historically oppressed groups' dialects, such as LGBTQIA+  dialects \cite{calder_2020}. Similarly to AAE, LGBTQIA+ dialects reclaim oppressive identity terms which in-group members use in a neutral or positive context. We also note that the study of biases towards AAE authors may be present in other NLP domains, such as sentiment analysis which is commonly used during hate detection's dataset creation. We hypothesize that our proposed HxEnsemble framework can be extended to these domains as we show its effectiveness in ensuring models are better able to represent the protected groups' target population.

\section{Related Work}
\label{related-works}
As machine learning systems are integrated as a standard tool of society, more researchers have begun to demonstrate how these systems learn and amplify biases from data \cite{Howard_UglyTruth_2017}. Bolukbasi et al. \cite{Bolukbasi16} demonstrated how widely pre-trained word embeddings trained on Google News articles capture sexist stereotypes. Similarly, Buolamwini and Gebru \cite{buolamwini18a} demonstrated how commercial gender ML systems perpetrate colorism by misclassifying dark-skinned women at higher rates than lighter-skinned individuals. In recent years, interest in hate-speech detection has grown as a sub-topic of natural language processing research and more datasets have become available for researchers \cite{fortuna2018, waseem-hovy-2016-hateful,DavidsonWMW17,golbeck18,Founta2018,kennedy2018,de-gibert-etal-2018-hate}. 

Hate speech and toxic language classification are riddled with challenges due to a low agreement in annotation, complexity in what constitutes hate speech, and lack of expertise that is required to understand the social and cultural structures that underlay different types of bigotry \cite{fortuna2018}. Work by Dixon et al. \cite{dixon2018} attempted to reduce the issue of "false-positive" biases associated with identity terms by re-balancing the dataset, while Kennedy et al. \cite{kennedy-etal-2020-contextualizing} used a post-hoc explanation regularizer to encourage the classifiers to learn the context around hate speech rather than group-identifiers. Work by Park et al. \cite{park-etal-2018-reducing} attempts to reduce gender bias in the toxic-language datasets by utilizing debiased word embeddings, data augmentation to swap gender pronouns, and the support of a larger corpus. 

More recently, limitations with the toxic-language datasets have become a source of research in this domain. For example, Awal et al. \cite{awal2020} showed that semantically similar samples in hate and abusive language datasets have issues with label consistency. As Fortuna and Nunes \cite{fortuna2018} discuss, a consistent definition of hate speech does not exist. This subjectivity can introduce annotation biases as demonstrated by the graph-based approach developed by Wich et al. \cite{wich-etal-2020-investigating} that groups annotators in order to identify annotated biases. In another study, Wich et al. \cite{wich-etal-2020-impact} demonstrated that politically-biased abusive language datasets impair the performance of hate speech classifiers. Sap et al. \cite{sap-etal-2019-risk} revealed a high correlation between annotators' perception of toxic labels and tweets predicted to be AAE using Blodgett et al.'s \cite{blodgett-etal-2016-demographic} dialect-prediction model. Furthermore, by using AAE dialect as a proxy for race, they showed that by relabeling a sample of the dataset with racial priming, the annotation bias towards AAE authors was significantly reduced. In both Sap et al. and Davidson et al.'s research, they demonstrated how using the toxic language datasets ad-hoc machine learning models propagate and amplify the racial biases towards AAE speakers \cite{sap-etal-2019-risk,davidson-etal-2019-racial}. Similarly to racial priming, Patton et al.'s Contextual Analysis of Social Media (CASM) framework introduced techniques to address the introduction of biases during data annotation \cite{casm_patton2020}. The CASM technique utilizes a multi-step process to have the annotators examine the cultural and contextualization of the data, as a result the authors observed improved classification results with reduction in annotation biases.

Xia et al. \cite{xia-etal-2020-demoting} and Zhou et al. \cite{zhou-etal-2021-challenges} have proposed approaches to minimize the racial biases that are propagated from these datasets. Xia et al. introduced an adversarial model architecture to reduce the false-positive rates of AAE while reducing the impact on classifier performance. Zhou et al. evaluated a pre-processing and in-processing debiasing techniques and introduced an experiment of relabeling the dataset by translating the AAE sample to SAE, in order to have the same toxicity label. Our work builds on these findings and adds to the state of knowledge by evaluating new debiasing techniques with additional fairness metrics, demonstrating challenges in using a larger corpus for bias-mitigation in low-resource contexts, and introducing an ensemble framework that increases fairness while minimizing classification degradation.

\section{Conclusion}
In this work, we proposed and evaluated an ensemble framework that leveraged a general toxic language classifier, dialect estimation model, and a specialized AAE classifier to reduce the racial biases in hate and toxicity detection datasets. We evaluated the HxEnsemble, two bias-mitigation algorithms, and common machine-learning classifiers using several fairness metrics and datasets that provided insights into how these models learn and propagate the annotation biases in the underlying datasets. Experiments conducted revealed that across all datasets, classifiers had higher FPR and  lower FNR for AAE instances than the SAE instances. Additionally, both favorable and unfavorable prediction biases exist against AAE authors, where the disparate impact score for non-toxic predictions is heavily biased against AAE authors, and predictions for toxic is heavily biased towards AAE authors. Although the data biases are propagated to the models, both our HxEnsemble and the explanation regularization bias-remediation techniques were able to mitigate some of the racial biases with minimal impact on classifier performance.

Using a thorough error analysis, we noted that the challenges in debiasing these datasets resulted from a large portion of misclassified AAE samples. We also presented characteristics of AAE samples that the HxEnsemble framework struggled with, which can further motivate future research on debiasing hate detection on AAE texts. Future usage of our proposed framework is extensible to other low-resource and biased domains, where it can be combined with other bias-mitigation techniques. Lastly, we demonstrated the effects of label consistency issues on classifier performances with two thresholds of dataset aggregation. We call for future work to create a new toxic language dataset that has AAE samples labeled by in-group annotators, has cultural training materials available, and/or adds racial priming or utilizes CASM to urge annotators to consider the cultural context of the tweet.

\begin{acks}
The authors thank the anonymous reviewers for the helpful feedback. This research was partially supported by a grant funded by Cisco Systems, Inc.
\end{acks}

\bibliographystyle{ACM-Reference-Format}
\bibliography{references}

\appendix

\section{Appendix}
\subsection{Fairness Metrics for Stricter AAE Prediction}
\label{app-fair08}
In the following tables we show the fairness metrics for the classifiers on each dataset using the threshold of $\text{Pr}(\text{AAE}\geq0.8)$ for texts belonging to the AAE dialect using Blodgett et al.'s model \cite{blodgett-etal-2016-demographic}. Please see \S\ref{sec:data-prep} for more information. 

A score of $0$ is used to indicate a missing metric or a division error. For example, in DWMW17 there are no mis-classified non-toxic AAE instances in the dataset, hence a FPR of $0$.

\begin{table*}[h]
\caption{DWMW17 Fairness Metrics for $\text{Pr}(\text{AAE}\geq0.8)$ [$n_{\text{AAE}} = 74, n_{\text{AAE}} = 2396$]}
\label{tab:dwmw1708}
\begin{tabular}{@{}lllllll@{}}
\toprule
Task Name  & $DI_{fav}$          & $DI_{unfav}$        & $FNR_{AAE}$         & $FNR_{SAE}$         & $FPR_{AAE}$         & $FPR_{SAE}$         \\ \midrule
N-Gram     & $0.000_{\pm 0.000}$ & $1.187_{\pm 0.002}$ & $0.000_{\pm 0.000}$ & $0.022_{\pm 0.001}$ & $0.000_{\pm 0.000}$ & $0.209_{\pm 0.004}$ \\
TF-IDF     & $0.220_{\pm 0.076}$ & $1.079_{\pm 0.007}$ & $0.020_{\pm 0.007}$ & $0.029_{\pm 0.000}$ & $0.000_{\pm 0.000}$ & $0.612_{\pm 0.003}$ \\
GloVe      & $0.000_{\pm 0.000}$ & $1.169_{\pm 0.020}$ & $0.000_{\pm 0.000}$ & $0.052_{\pm 0.009}$ & $0.000_{\pm 0.000}$ & $0.424_{\pm 0.045}$ \\
BERT       & $0.055_{\pm 0.038}$ & $1.196_{\pm 0.010}$ & $0.009_{\pm 0.007}$ & $0.019_{\pm 0.002}$ & $0.000_{\pm 0.000}$ & $0.114_{\pm 0.014}$ \\
BERT+OC    & $0.000_{\pm 0.000}$ & $1.222_{\pm 0.007}$ & $0.000_{\pm 0.000}$ & $0.024_{\pm 0.003}$ & $0.000_{\pm 0.000}$ & $0.086_{\pm 0.012}$ \\
BERT+SOC   & $0.000_{\pm 0.000}$ & $1.222_{\pm 0.004}$ & $0.000_{\pm 0.000}$ & $0.023_{\pm 0.001}$ & $0.000_{\pm 0.000}$ & $0.082_{\pm 0.009}$ \\
BERT+MD    & $0.026_{\pm 0.061}$ & $1.090_{\pm 0.098}$ & $0.005_{\pm 0.013}$ & $0.013_{\pm 0.022}$ & $0.000_{\pm 0.000}$ & $0.606_{\pm 0.416}$ \\
HxEnsemble & $0.056_{\pm 0.039}$ & $1.192_{\pm 0.008}$ & $0.009_{\pm 0.007}$ & $0.018_{\pm 0.001}$ & $0.000_{\pm 0.000}$ & $0.129_{\pm 0.008}$ \\ \bottomrule
\end{tabular}
\end{table*}

\begin{table*}[h]
\caption{FDCL18 Fairness Metrics for $\text{Pr}(\text{AAE}\geq0.8)$ [$n_{\text{AAE}} = 7, n_{\text{SAE}} = 9079$]}
\label{tab:fdcl1808}
\begin{tabular}{@{}lllllll@{}}
\toprule
Task Name  & $DI_{fav}$            & $DI_{unfav}$        & $FNR_{AAE}$         & $FNR_{SAE}$         & $FPR_{AAE}$         & $FPR_{SAE}$         \\ \midrule
N-Gram     & $0.193_{\pm 0.000}$   & $3.313_{\pm 0.015}$ & $0.000_{\pm 0.000}$ & $0.143_{\pm 0.002}$ & $0.000_{\pm 0.000}$ & $0.036_{\pm 0.001}$ \\
TF-IDF     & $0.549_{\pm 0.001}$   & $2.597_{\pm 0.010}$ & $0.333_{\pm 0.000}$ & $0.286_{\pm 0.002}$ & $0.000_{\pm 0.000}$ & $0.036_{\pm 0.000}$ \\
GloVe      & $0.191_{\pm 0.001}$   & $3.377_{\pm 0.034}$ & $0.000_{\pm 0.000}$ & $0.217_{\pm 0.005}$ & $0.000_{\pm 0.000}$ & $0.057_{\pm 0.002}$ \\
BERT       & $0.198_{\pm 0.000}$   & $3.090_{\pm 0.016}$ & $0.000_{\pm 0.000}$ & $0.094_{\pm 0.003}$ & $0.000_{\pm 0.000}$ & $0.044_{\pm 0.001}$ \\
BERT+OC    & $0.196_{\pm 0.001}$   & $3.152_{\pm 0.023}$ & $0.000_{\pm 0.000}$ & $0.104_{\pm 0.003}$ & $0.000_{\pm 0.000}$ & $0.040_{\pm 0.002}$ \\
BERT+SOC   & $0.176_{\pm 0.062}$   & $3.201_{\pm 0.132}$ & $0.000_{\pm 0.000}$ & $0.103_{\pm 0.005}$ & $0.100_{\pm 0.316}$ & $0.040_{\pm 0.003}$ \\
BERT+MD    & $0.582_{\pm 0.445}$   & $1.531_{\pm 1.647}$ & $0.500_{\pm 0.527}$ & $0.554_{\pm 0.470}$ & $0.100_{\pm 0.316}$ & $0.035_{\pm 0.045}$ \\
HxEnsemble & $0.197_{\pm   0.000}$ & $3.119_{\pm 0.019}$ & $0.000_{\pm 0.000}$ & $0.098_{\pm 0.004}$ & $0.000_{\pm 0.000}$ & $0.042_{\pm 0.001}$ \\ \bottomrule
\end{tabular}
\end{table*}

\begin{table*}[h]
\caption{Toxic Fairness Metrics for $\text{Pr}(\text{AAE}\geq0.8)$ [$n_{\text{AAE}} = 70, n_{\text{SAE}} = 13376$]}
\label{tab:toxic08}
\begin{tabular}{@{}lllllll@{}}
\toprule
Task Name  & $DI_{fav}$            & $DI_{unfav}$        & $FNR_{AAE}$         & $FNR_{SAE}$         & $FPR_{AAE}$         & $FPR_{SAE}$         \\ \midrule
N-Gram     & $0.022_{\pm   0.000}$ & $2.774_{\pm 0.013}$ & $0.000_{\pm 0.000}$ & $0.172_{\pm 0.002}$ & $0.500_{\pm 0.000}$ & $0.055_{\pm 0.001}$ \\
TF-IDF     & $0.144_{\pm 0.006}$   & $2.874_{\pm 0.024}$ & $0.101_{\pm 0.005}$ & $0.297_{\pm 0.004}$ & $1.000_{\pm 0.000}$ & $0.066_{\pm 0.002}$ \\
GloVe      & $0.056_{\pm 0.012}$   & $2.687_{\pm 0.025}$ & $0.037_{\pm 0.008}$ & $0.245_{\pm 0.002}$ & $1.000_{\pm 0.000}$ & $0.107_{\pm 0.002}$ \\
BERT       & $0.043_{\pm 0.007}$   & $2.688_{\pm 0.013}$ & $0.000_{\pm 0.000}$ & $0.142_{\pm 0.001}$ & $0.050_{\pm 0.158}$ & $0.046_{\pm 0.001}$ \\
BERT+OC    & $0.022_{\pm 0.015}$   & $2.726_{\pm 0.041}$ & $0.000_{\pm 0.000}$ & $0.150_{\pm 0.005}$ & $0.500_{\pm 0.333}$ & $0.051_{\pm 0.003}$ \\
BERT+SOC   & $0.034_{\pm 0.012}$   & $2.695_{\pm 0.023}$ & $0.000_{\pm 0.000}$ & $0.148_{\pm 0.004}$ & $0.250_{\pm 0.264}$ & $0.052_{\pm 0.003}$ \\
BERT+MD    & $0.722_{\pm 0.503}$   & $0.397_{\pm 0.515}$ & $0.644_{\pm 0.477}$ & $0.629_{\pm 0.486}$ & $0.300_{\pm 0.483}$ & $0.343_{\pm 0.457}$ \\
HxEnsemble & $0.056_{\pm   0.012}$ & $2.627_{\pm 0.032}$ & $0.007_{\pm 0.008}$ & $0.138_{\pm 0.003}$ & $0.000_{\pm 0.000}$ & $0.053_{\pm 0.003}$ \\ \bottomrule
\end{tabular}
\end{table*}

\begin{table*}[h]
\caption{Hate Fairness Metrics for $\text{Pr}(\text{AAE}\geq0.8)$ [$n_{\text{AAE}} = 66, n_{\text{SAE}} = 12323$]}
\label{tab:hate08}
\begin{tabular}{@{}lllllll@{}}
\toprule
Task Name  & $DI_{fav}$            & $DI_{unfav}$        & $FNR_{AAE}$         & $FNR_{SAE}$         & $FPR_{AAE}$         & $FPR_{SAE}$         \\ \midrule
N-Gram     & $0.980_{\pm   0.007}$ & $1.631_{\pm 0.228}$ & $0.800_{\pm 0.000}$ & $0.735_{\pm 0.008}$ & $0.038_{\pm 0.008}$ & $0.013_{\pm 0.001}$ \\
TF-IDF     & $0.968_{\pm 0.000}$   & $3.330_{\pm 0.081}$ & $0.800_{\pm 0.000}$ & $0.860_{\pm 0.002}$ & $0.033_{\pm 0.000}$ & $0.004_{\pm 0.000}$ \\
GloVe      & $1.019_{\pm 0.008}$   & $0.000_{\pm 0.000}$ & $1.000_{\pm 0.000}$ & $0.884_{\pm 0.031}$ & $0.000_{\pm 0.000}$ & $0.011_{\pm 0.005}$ \\
BERT       & $0.991_{\pm 0.015}$   & $1.154_{\pm 0.258}$ & $0.680_{\pm 0.103}$ & $0.533_{\pm 0.008}$ & $0.043_{\pm 0.011}$ & $0.024_{\pm 0.001}$ \\
BERT+OC    & $0.969_{\pm 0.022}$   & $2.002_{\pm 0.745}$ & $0.860_{\pm 0.165}$ & $0.722_{\pm 0.037}$ & $0.054_{\pm 0.019}$ & $0.011_{\pm 0.004}$ \\
BERT+SOC   & $0.956_{\pm 0.039}$   & $2.317_{\pm 1.316}$ & $0.720_{\pm 0.193}$ & $0.707_{\pm 0.027}$ & $0.059_{\pm 0.029}$ & $0.014_{\pm 0.003}$ \\
BERT+MD    & $0.400_{\pm 0.516}$   & $0.600_{\pm 0.516}$ & $0.400_{\pm 0.516}$ & $0.400_{\pm 0.516}$ & $0.600_{\pm 0.516}$ & $0.600_{\pm 0.516}$ \\
HxEnsemble & $1.042_{\pm   0.011}$ & $0.210_{\pm 0.204}$ & $0.960_{\pm 0.126}$ & $0.551_{\pm 0.009}$ & $0.008_{\pm 0.009}$ & $0.021_{\pm 0.001}$ \\ \bottomrule
\end{tabular}
\end{table*}

\end{document}